\documentclass[letterpaper]{article} 
\usepackage{aaai23}  
\usepackage{times}  
\usepackage{helvet}  
\usepackage{courier}  
\usepackage[hyphens]{url}  
\usepackage{graphicx} 
\usepackage{natbib}  
\usepackage{caption} 
\usepackage[linesnumbered,ruled,vlined]{algorithm2e}
\usepackage{fancyvrb}
\usepackage{mdframed}
\usepackage{tabularx}
\usepackage{multirow}
\usepackage{amsmath}
\usepackage{tikz}
\usepackage{pgfplots}
\usepackage{datatool}
\usepackage{pgfplotstable}
\usepackage{rotating} 
\usepackage{cprotect}  

\usepackage{newfloat}
\usepackage{listings}
\usepackage{bm}
\DeclareCaptionStyle{ruled}{labelfont=normalfont,labelsep=colon,strut=off} 
\usepackage{subcaption} 
\newcolumntype{s}{>{\hsize=0.8\hsize}X}
\newcolumntype{B}{>{\hsize=1.2\hsize}X}

\usepackage{fancyhdr}
\title{Design and Implementation of Automatic Assisted Aiming System \\For Robomaster EP Based on YOLOv5}

\author {
    Junjia Qin,\textsuperscript{\rm 1}
    Kangli Xu\textsuperscript{\rm 1}
}
\affiliations {
    \textsuperscript{\rm 1} The High School Affiliated to Minzu University of China
}

\begin{document}

\maketitle

\begin{abstract}
In the crucial stages of the Robomaster Youth Championship, the Robomaster EP Robot must operate exclusively on autonomous algorithms to remain competitive. Target recognition and automatic assisted aiming are indispensable for the EP robot.  
In this study, we use YOLOv5 for multi-object detection to identify the Robomaster EP Robot and its armor. Additionally, we integrate the DeepSORT algorithm for vehicle identification and tracking. As a result, we introduce a refined YOLOv5-based system that allows the robot to recognize and aim at multiple targets simultaneously.  
To ensure precise tracking, we use a PID controller with Feedforward Enhancement and an FIR controller paired with a Kalman filter. This setup enables quick gimbal movement towards the target and predicts its next position, optimizing potential damage during motion. 
Our proposed system enhances the robot's accuracy in targeting armor, improving its competitive performance. 

\end{abstract}

\section{Introduction}

\quad 
In the research and development of the target identification and Automatic Assisted Aiming System (AAA-System) for the RoboMaster EP Robot, mastering fundamental knowledge across multiple disciplines and integrating key technologies from various fields are imperative. This paper focuses on the infantry robot equipped with a three-axis gimbal and a chassis based on the Mecanum wheel. This robot scores points in competitions by firing water pellets at the armor plates of opponents, playing a pivotal role in determining the contest's outcome. Our study of the AAA-System centers on this infantry robot. 

YOLOv5, an advanced detection model, excels in managing high-density image data in robotic tactical battles, particularly when classifying overlapping areas. Its structure and efficiency are apt for real-time processing, crucial in rapid robotic encounters. CUDA's acceleration is pivotal in the realm of embedded systems like the NVIDIA Jetson series, enhancing the optimization of deep learning models. Notably, YOLOv5's architecture is amenable to conversion into a TensorRT model, benefiting from CUDA's increased inference speed. Hence, this paper employs YOLOv5 as our automatic assisted aiming system's detection component, capitalizing on the Jetson board's computational capabilities with CUDA acceleration. 

We employ DeepSORT, an object tracking algorithm. This algorithm can automatically learn and assign unique IDs to objects, preserving them even amidst challenges like occlusions or changing camera angles. Its computational efficiency deems it apt for real-time robot competitions. Additionally, the Hungarian algorithm aids in real-time strike data tracking, fine-tuning the targeting sequence. 

To accurately predict target trajectories, especially in high-dynamic settings, we utilize the Kalman filter alongside PID with Feedforward Enhancement and FIR controllers. The Kalman filter excels in optimizing state estimates for systems prone to random disturbances. The PID controller, with its Feedforward Enhancement, offers timely corrections by gauging the variance between the desired and actual system outputs. Furthermore, the FIR controller finetunes these corrections, ensuring system stability against rapid target movements or unforeseen disturbances. 

\textbf{\textsc{Our key findings are}}
1) YOLOv5 detection model is aptly suited for the robot's tactical combat due to its high-density image information processing, especially in areas of overlapping classifications;
2) The integration of the DeepSORT algorithm with the Hungarian method offers real-time target tracking and determination of the optimal shooting sequence, further augmented by the use of Kalman filters and PID with FIR controllers for trajectory prediction.
3) Use strided convolution instead of MaxPooling for small object optimization in YOLOv5.
They facilitate gimbal tracking, when YOLOv5 experiences recognition delays and the robot moves swiftly, ultimately Maximizing Competition Scores.

\section{Background and Related Work}

\quad 
DJI RoboMaster is a premier global Youth Robotics Competition. In the contest, competing teams operate robots they've either developed autonomously or modified. These robots participate in \textbf{4 vs 4} tactical shooting match within a designated arena. Teams maneuver their robots to fire water pellets, with the aim of hitting opposing robots or their bases. During the course of the competition, the team that first destroys the opponent's base emerges victorious. However, if neither base is destroyed by the end of the match, the team with more remaining base health is declared the winner. The infantry robot serves as the primary combat force of each team, responsible for firepower output. This requires the infantry robot to perform tasks such as visual recognition. Primarily, the infantry robot is outfitted with 2 control systems: \textit{Chassis and Gimbal}. \textbf{Our focus in this study is on visual algorithm tracking, centered on the Three-Axis control of the gimbal.} \newline 

\noindent \textbf{A.} \textit{Evolution of Visual Algorithms and Comparison with Other Methods} \newline 

Traditional visual recognition techniques, such as the Harris corner detection algorithm and \textit{OpenCV Brute-Force matching}, often fall short in providing adequate accuracy in real-world scenarios. Subsequent detection models based on deep learning, such as \textit{Fast R-CNN}, \textit{Faster R-CNN}, \textit{SSD}, enhance the precision of target recognition. However, they often lack computational efficiency, especially when processing high-density image data in scenarios like robot combat.

The YOLO series saw several iterations, with each version enhancing its predecessors in various aspects. YOLOv5, emerged as a pinnacle in this series. Compared to the earlier versions, YOLOv5 showcases: 
\begin{enumerate}
    \item \textbf{Improved accuracy}, The enhanced architecture and fine-tuning capabilities make YOLOv5 one of the most accurate detection models available, outperforming even some of the previously discussed models.

    \item \textbf{Efficient computation}, Despite its heightened accuracy, YOLOv5 does not compromise on speed. Its optimized structure ensures rapid detection, making it suitable for real-time applications like robot combat.

    \item \textbf{Scalability}, YOLOv5 provides multiple model sizes, allowing users to make trade-offs between speed and accuracy based on their specific application requirements. 
\end{enumerate}

YOLOv5 stands out as an optimal convolutional deep learning detection model, particularly for applications demanding both high-speed processing and precision. \newline 

\noindent \textbf{B.} \textit{Predictive and Tracking Robots} \newline

We utilize the tracking capability of DeepSORT (Simple Online and Realtime Tracking with a Deep Association Metric) to label the Armors and record which ones have been hit, all while tracking the movement trajectory of each armor. It uses dynamic programming to decide which armors should be prioritized for hitting. When seeking a target to hit, there is a need to find the optimal solution. The optimal solution for a given calculation often encompasses optimal solutions for its sub-problems. 

DeepSORT is a target tracking algorithm based on deep learning, used for tracking one or multiple targets. It employs a re-identification network for feature extraction, preserving distinct feature information between targets to reduce identity switches after occlusion. It also adopts the Kalman Filtering Algorithm to manage trajectories and uses the cascade matching algorithm, IOU matching algorithm, and the Hungarian algorithm for matching. 

\begin{figure*}[t]
	\centering
	\captionsetup{justification=centering} 
	\includegraphics[width=0.8\textwidth]{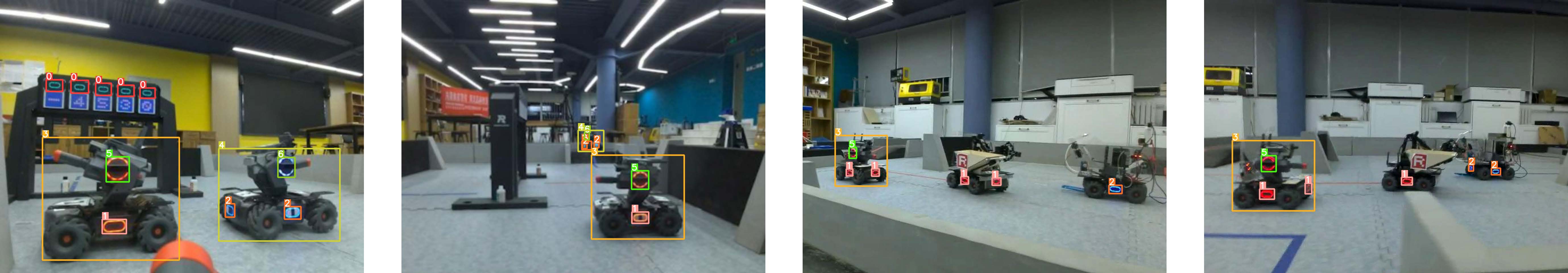}
	\caption{Recognizable objects.}
	\label{fig:image_targets}
\end{figure*}

The advantage of this algorithm is that it can track multiple targets and continue tracking them even when they are occluded. \newline 

\noindent \textbf{C.} \textit{Control and Filtering Mechanisms in Robotic Combat} \newline

In the realm of robotic combat, simply tracking a target is insufficient. Equally crucial is the capability to predict the target's trajectory, ensuring the most effective shooting sequence.

Many robotic systems suffer from sensor noise, leading to imprecise readings and, subsequently, flawed decision-making. However, FIR filters, linear filters engineered to act based solely on input signals, stand out as a solution. They effectively reduce noise without introducing phase delay, ensuring that combat robots' gimbals obtain clearer data for more accurate responses.

In robotic combat systems, there's a clear need for methodologies capable of predicting the state (Position, Velocity) of dynamic targets. Among the myriad of solutions, Kalman filters, imbued with a recursive algorithm, emerge preeminent. Their strength lies in skillfully estimating the state of a process using temporal observational data. The adoption of Kalman filters furnishes combat robots with an enhanced apparatus for accurately foreseeing adversary maneuvers, thereby honing target acquisition and augmenting shooting precision. 

In previous research on robotic combat, which requires a high degree of precision, speed, and adaptability, there was a heavy reliance on basic control algorithms and filtering mechanisms. While these initial robotic configurations were operational and could engage in competition, they encountered a number of challenges:

\begin{enumerate}
    \item \textbf{Inconsistent Responses}: Without the aid of advanced controllers, robots often demonstrated unpredictable reactions to environmental alterations. This inconsistency in responses engendered erratic movement patterns, leading to disjointed and ineffective combat maneuvers.
    
    \item \textbf{Noise and Disturbance}: Early robotic systems were devoid of effective means to mitigate sensor noise or external disturbances, leaving them susceptible to misinterpretations and malfunctions.

    \item \textbf{Limited Processing Capabilities}: Early robotic systems were often constrained by their processing capabilities, which hampered real-time decision-making and adaptive responses in dynamic combat scenarios. The inability to process complex environmental data swiftly resulted in delayed reactions and, consequently, sub-optimal combat strategies.
\end{enumerate}
    
The incorporation of PID controllers, FIR filters, and Kalman filters into robotic combat systems addresses the historical challenges confronted by their predecessors. These technological advancements significantly enhance the precision, speed, and adaptability of combat robots, marking the advent of a new epoch in robotic combat. 

\subsection{Programming Task}
The overarching approach of this research integrates various interrelated components. These components encompass the integration of design objectives and functional requirements, the combination of physical modeling and virtual simulation, the alignment of prototype creation with empirical validation, and the synchronization of movement choreography with motion characteristics. In our quest to enable robots to accurately identify and track multiple targets, we drew insights from the visual development process. The key tasks undertaken in this regard are: 

\begin{enumerate}
    \item Investigating the features of the YOLOv5 detection model and selecting the one that best fits our requirements for YOLOv5n-v6. 
    \item Collecting the necessary images. To ensure accurate annotations, manual labeling was employed.
    \item Setting up the training environment and initiating the model training process.
    \item During training, it was observed that there was a deviation between our anticipated results and the actual outcomes. This prompted us to fine-tune the discrepancies encountered during training and optimize the model accordingly.
    \item Selecting the Optimal Development Board, Detection Hardware. 
    \item Integrating YOLOv5 with DeepSORT to achieve target tracking by ID, even under dynamic conditions or when the target becomes occluded.
    \item Optimizing Target Striking, Fitting the falling scenario of water pellets. 
    \item Incorporating a PID controller with Feedforward Enhancement and FIR filter to precisely predict and control the movement of the turret. 
\end{enumerate} 

\section{Experiment 1: End-to-End Model Development: From Dataset Construction to Deployment}

\subsection{Dataset Construction}

The \textit{Robomaster competition arena} possesses a distinctive rectangular shape, measuring approximately \textit{7 meters} in length and \textit{5 meters} in width. This arena has a varied terrain, featuring gradients such as \textit{15°} and \textit{30°}. Additionally, the field is divided into specific zones, among which are the base information area and the energy mechanism section. 

To support progressive research endeavors and enable effective model training, we meticulously developed the AAAS-2021 dataset rooted in the aforementioned competition field attributes. 

A corpus of over 20K images was collected under standardized lighting conditions, sourced from an array of positions and perspectives within the competition arena, guaranteeing exhaustive representation. In our pursuit to emulate varied environmental luminosities, an auxiliary set of 8,000 photographs was gathered from assorted vantage points, culminating in a comprehensive dataset of 28,000 images.  

A rigorous annotation process was applied to all the images in accordance with established benchmarks. The collection was subsequently partitioned in a \textbf{6:2:2} ratio, allocating \textbf{16,000} images for Train, \textbf{4,200} images for Valid, and the remaining \textbf{4,000} for Test. \newline 

\textit{Armor} refers to the oval-shaped LED target blocks on Robomaster EP. The mechanism for hit detection works by firing water pellets at the Armor, which then emits a sound at a fixed frequency to determine if it has been struck. \textit{Army} refers to the infantry robots of Robomaster EP. \textit{Ear} denotes the LED circular lights on the Robomaster's three-dimensional turret used for identifying the current Yaw axis posture; these lights are non-targetable. Before the start of the competition, the vehicle will change the LED lights it carries (located on the Armor and Ear) to one of two colors: either \textcolor{red}{Red} or \textcolor{blue}{Blue}. When deactivated or dead in competition, the LED lights will go off. 

We pre-classified the entire set of recognizable objects, listing a total of 11 targets: 0) \textbf{\textcolor{cyan}{CyanArmor}}, 1) \textbf{\textcolor{red}{RedArmor}}, 2) \textbf{\textcolor{blue}{BlueArmor}}, 3) \textbf{\textcolor{red}{RedArmy}}, 4) \textbf{\textcolor{blue}{BlueArmy}}, 5) \textbf{\textcolor{red}{RedEar}}, 6) \textbf{\textcolor{blue}{BlueEar}}, 7) \textbf{\textcolor{red}{RedBase}}, 8) \textbf{\textcolor{blue}{BlueBase}}, 9) \textbf{DeadArmor}, 10) \textbf{DeadArmy}.\newline
\noindent \textbf{The annotations played a crucial role in accurately identifying objects in complex image environments during robot competitions.} \newline 

\subsection{Pre-Trained Configuration}
The configuration used for training is:
\begin{itemize}
    \item \textbf{CPU:} Intel Core i7 9700K
    \item \textbf{Memory:} 128 GB DDR4 RAM
    \item \textbf{GPU:} Nvidia RTX 2080
    \item \textbf{Operating System:} Windows 10
    \item \textbf{Programming Language:} Python 3.9
    \item \textbf{Deep Learning Framework:} Pytorch 1.10
\end{itemize}

\subsection{Train Model}

\begin{table*}
    \small
    \centering
    \renewcommand{\arraystretch}{1.0}
    \begin{tabular}{|c|c|c|c|c|c|c|c|c|}
    \hline
    $ \textbf{Model Types} $ & $ \textbf{epoch} $ & $ \textbf{box\_loss} $ & $ \textbf{obj\_loss} $ & $ \textbf{cls\_loss} $ & $ \textbf{precision} $ & $ \textbf{recall} $ & $ \textbf{mAP}_{0.5} $ & $ \textbf{mAP}_{0.5:0.95} $ \\
    \hline
    Y5n & 200 & 0.024 & 0.018 & 0.0044 & 0.933 & 0.942 & 0.963 & 0.735 \\
    Y5n-SC & 200 & 0.023 & 0.017 & 0.0042 & 0.934 & 0.941 & 0.963 & 0.735 \\
    Y5n & 300 & 0.022 & 0.017 & 0.0041 & 0.933 & 0.941 & \textbf{0.964} & 0.736 \\
    Y5n-SC & 300 & 0.022 & 0.016 & 0.0034 & 0.935 & \textbf{0.949} & 0.963 & 0.735 \\
    Y5n & 400 & \textbf{0.021} & 0.0164 & \textbf{0.0029} & 0.928 & 0.945 & 0.963 & 0.734 \\
    Y5n-SC* & 400 & \textbf{0.021} & \textbf{0.0162} & 0.0031 & \textbf{0.935} & 0.946 & \textbf{0.964} & \textbf{0.735} \\
    Y5n & 500 & 0.023 & 0.017 & 0.0035 & 0.926 & 0.937 & 0.952 & 0.732 \\
    Y5n-SC & 500 & 0.024 & 0.017 & 0.0039 & 0.927 & 0.939 & 0.948 & 0.722 \\
    \hline
    \end{tabular}
    \caption{\centering Results for different metrics across epochs.}
    \label{tab:results-epochs}
\end{table*}

When training deep learning models, the loss function serves as an indicator to gauge the discrepancy between the model's predictions and the actual outcomes. In YOLOv5, we utilize \textit{yolov5n.pt} for training. We employ three distinct loss functions: \textit{box\_loss}, \textit{obj\_loss}, and \textit{cls\_loss}. Additionally, we reference \textit{mAP\_0.5} and \textit{mAP\_0.5:0.95}.


\subsubsection{box\_loss}
This loss function evaluates the prediction error concerning the bounding box's position. It's derived by squaring the differences between the predicted and actual coordinates, lengths, and widths of the bounding box, followed by averaging the results.


\subsubsection{obj\_loss}
This loss function assesses the prediction error for the object's presence within the bounding box. It's computed by summing the cross-entropy between the predicted and actual probabilities of the object's presence, then averaging the results.


\subsubsection{cls\_loss}
This loss function evaluates the prediction error for the object's category within the bounding box. It's derived by summing the cross-entropy between the predicted and actual category probabilities, followed by averaging the results.


\subsubsection{metrics\_precision} 

Precision, in binary classification, measures the accuracy of positive predictions. It is defined as the ratio of true positives (TP) to the sum of true positives and false positives (FP):

\begin{equation}
\text{Precision} = \frac{TP}{TP + FP}
\end{equation}

High precision indicates a lower number of false positives. However, it does not necessarily reflect the capture of all actual positives; this aspect is addressed by recall.

\subsubsection{metrics\_recall} 

Recall, also known as sensitivity, measures the model's capability to detect relevant instances. In binary classification, it is the ratio of true positives (TP) to the sum of true positives and false negatives (FN):

\begin{equation}
\text{Recall} = \frac{TP}{TP + FN}
\end{equation}

While a high recall signifies that the model identifies most of the actual positives, a holistic evaluation requires balancing it with precision.

\subsubsection{mAP (mean Average Precision)} 

The mean Average Precision (mAP) serves as a crucial metric for evaluating object detection models, as it summarizes the precision-recall curve across all dataset classes. For each class, the Average Precision (AP) is derived from the area beneath its precision-recall curve, representing the mean precision at varying recall levels. The mAP is then determined by averaging the APs for all classes:

\begin{equation}
\text{mAP} = \frac{1}{N} \sum_{i=1}^{N} AP_i
\end{equation}

\( N \) is the total number of classes and \( AP_i \) is the average precision for class \( i \). 

A superior mAP score suggests that the object detection model excels in both recall (detecting most objects) and precision (ensuring the detected objects are predominantly relevant). This metric is valuable because it encompasses the model's performance across all classes and thresholds, offering a comprehensive view of its effectiveness.

\subsection{YOLOv5n Model Optimization}

In practical recognition scenarios, the \textit{Armor} is represented as a detection box with a \textbf{small resolution (resolution smaller than 32x32).} \\ 

In the original version of YOLOv5n, multiple MaxPooling computations are employed. During the training phase, it was observed that the traditional \textbf{MaxPooling layers consumed a considerable amount of computational time}, thereby reducing the efficiency of model training. Consequently, we opted for Strided-Convolution as an alternative to the MaxPooling layer. Strided-Convolution achieves the downsampling effect of the MaxPooling layer by adjusting the stride of the convolution kernel. Moreover, as it does not entail additional computations, it effectively saves time. \newline 

Strided-Convolution is ideal for lightweight models as it can downsample without increasing the number of parameters, thereby reducing the computational load of the model. Downsampling of the input can be achieved by setting the Stride parameter, where a larger Stride results in more aggressive downsampling. For each input image, the Strided-Convolution outputs a feature map of the same size as the input. This allows for downsampling while maintaining the size of the feature map, making it possible to use Strided-Convolution in place of the 3x3 MaxPooling layer by adjusting the stride parameter. 

By setting the stride value to 2, the input can be effectively downsampled to half of its original size, thereby reducing the spatial dimensions of the output. For instance, if our input has dimensions of 28x28x32 and a convolution layer with a stride of 2 is applied over it, the output dimensions would be 14x14x32. This can serve as a useful alternative to shrinking the spatial dimensions of the input, as it allows us to retain more spatial information from the original input. Using Strided-Convolution is computationally more efficient, as it eliminates the need for additional MaxPool layers in the network. 

Based on the experimental results, we observed that as the number of epochs increases, the performance of the model improves. We found that the YOLOv5n model's performance is optimal when the epochs reach around 400. At this point, the values of \textit{box\_loss}, \textit{obj\_loss}, and \textit{cls\_loss} are relatively low, and the values of \textit{precision}, \textit{recall}, and \textit{mAP} are high. We believe that after more than \textbf{400 epochs}, the model's accuracy shows signs of overfitting. In particular, there is a significant decline in \textit{recall}. Overfitting often leads to a decrease in the model's generalization ability, so care should be taken to avoid overfitting when training models. 

In Table \ref{tab:results-epochs}, the Original Version with MaxPooling is referred to as the \textbf{Y5n} Model Type, and the one using Strided-Convolution in place of the 3x3 MaxPooling layer as the \textbf{Y5n-SC} Model Types. We observed that under the same epochs, the \textbf{Y5n-SC} even showcased better numerical performance than \textbf{Y5n}. Thus, \textbf{we believe that adopting Strided-Convolution as a replacement for the MaxPool layer is an effective optimization method}, capable of enhancing the model's training efficiency and being suitable for lightweight models. 

\subsection{Model Deploy}

\begin{table*}
    \small 
    \centering
    \renewcommand{\arraystretch}{1.0}
    \begin{tabular}{|c|c|c|c|c|c|c|c|c|}
    \hline 
    Device & Model Types & Avg ms/100 & Avg ms/1k & ms at 320x320 & $AP_{50}$ & $AP_{small}$ \\
    \hline
    Nvidia RTX 2080 (CUDA) & Y5n & 9.41ms & 9.39ms & 4ms & 0.951 & 0.681 \\
    Nvidia RTX 2080 (CUDA)* & Y5n-SC & \textbf{8.84ms} & \textbf{8.82ms} & \textbf{4ms} & 0.951 & \textbf{0.682} \\
    Intel i7-9700K (CPU) & Y5n & 45.4ms & 45.6ms & 25ms & \textbf{0.958} & 0.672 \\
    Intel i7-9700K (CPU) & Y5n-SC & 42.1ms & 42.2ms & 25ms & 0.955 & 0.671 \\
    Jetson Nano (CUDA) & Y5n & 35.22ms & 36.64ms & 15ms & \textbf{0.958} & 0.675 \\
    Jetson Nano (CUDA)* & Y5n-SC & \textbf{28.44ms} & \textbf{28.31ms} & \textbf{12ms} & 0.957 & \textbf{0.676} \\
    RK3588 (TPUs) & Y5n & 63.2ms & 63.1ms & 32ms & 0.945 & 0.672 \\
    RK3588 (TPUs) & Y5n-SC & 62.1ms & 61.9ms & 29ms & 0.946 & 0.671 \\
    \hline
    \end{tabular}
    \caption{\centering Detection Performance Metrics for Different Devices and Model Types.}
    \label{tab:deploy-detections}
\end{table*}

Using the YOLOv5 architecture and implemented in PyTorch, this framework exemplifies modern deep learning techniques. \textit{PyTorch}, a highly praised deep learning library, offers a dynamic computational graph that not only facilitates easier debugging but also provides a more intuitive platform for model design. Its intrinsic Python-centric architecture ensures seamless integration with various Python libraries and tools, which proves invaluable during the training phase. \textsc{PyTorch models can be enhanced and expedited through a multitude of acceleration techniques:} 

1) \textit{TensorRT} is a high-performance deep learning inference optimizer that can be used to accelerate PyTorch models on NVIDIA GPUs. 2) \textit{Neural Processing Units (NPUs)} are engineered to efficiently handle tasks associated with neural networks, NPUs emerge as a vital piece of hardware, streamlining computations and enhancing overall system performance. 

In our experiments, we considered multiple hardware configurations to gauge the best platform for deployment. We selected the \textit{Nvidia Jetson Nano}, \textit{Nvidia RTX 2080}, \textit{Rockchip RK3588}, and \textit{Intel i7 9700K CPU} as our testbeds. These devices were chosen based on their popularity and diverse range of capabilities, ensuring a comprehensive performance analysis. 

In Table \ref{tab:deploy-detections}, we utilized a test set distinct from the main dataset for evaluation.

All evaluations on the hardware were conducted using the fp32 model.
\begin{itemize}
	\item The term \textit{Avg ms/100} indicates the average time consumed for detecting a target 100 times.
	\item \textit{Avg ms/1k} stands for the average time taken for detecting a target 1000 times.
	\item \textit{ms at 320x320} is the detect time when the resolution is other than 640x640.
	\item $AP_{50}$ evaluates the accuracy of target localization. 
	\item $AP_{\textit{small}}$ assesses the model's performance in detecting small targets.
\end{itemize}

The CUDA-accelerated Jetson Nano clearly stood out as a superior choice. The Jetson Nano, with its dedicated architecture for AI and deep learning tasks, proved to be a highly efficient device for deploying our PyTorch models, outshining other contenders in terms of speed and resource utilization. We utilized TensorRTX for deploying deep learning models on NVIDIA GPUs. TensorRTX is compatible with multiple deep learning frameworks and can optimize models for efficient CUDA execution. 

While \textit{PyTorch} offers numerous advantages for training, selecting the appropriate deployment device is equally crucial. In our case, the CUDA-accelerated Jetson Nano demonstrated superior capabilities, marking it as an optimal choice for our deployment needs. 

\section{Experiment 2: Identifying Armor and Tracking Robots}

The information recognized by YOLOv5 comprises object ID, top-left corner, bottom-right corner, and confidence. 

We input the targets detected by YOLOv5 into the \textit{recv} array. The data captured by the array includes object ID, color, length, width, center coordinates, and more. Subsequently, we employ DeepSORT to assign tracking numbers to the array data and record the movement trajectories of each armor, indicating which ones have been hit. 

Identifying the optimal solution is imperative when searching for a target to hit. An optimal solution for a given problem inherently encompasses the optimal solutions for its sub-problems. This inherent characteristic is referred to as the 'optimal substructure property.' During the analysis of a problem's optimal substructure, systematic methods are employed. Leveraging the optimal substructure property of a problem allows for the iterative construction of the optimal solution for the entire problem using a bottom-up approach. 

\textbf{\textsc{We proposed our optimization strategy:}} 1) Employ DeepSORT to classify and store the pre-encoded IDs, log the current coordinates and timestamp, note the IMU state, and convert the ID into 3D information for mapping. 2) For each armor, we can log its position and size data and calculate its distance from the center point, allowing us to directly compare distances to determine which armor should be targeted. 3) Use the Kalman filter to predict the armor plate's location during the time interval of the YOLOv5 update cycle, and employ the FIR filter combined with PID control to maneuver the gimbal. 

\subsection{DeepSORT Tracking}
DeepSORT provides an integration of motion and appearance information, resulting in improved tracking performance. 

The prediction encompasses both position and velocity, giving an estimate of where the target might be in the subsequent frame. 

\begin{equation}
{\mathbf{X}_{k|k-1}} =  {\mathbf{F}_k} \cdot  {\mathbf{X}_{k-1|k-1}}
\end{equation}

\begin{equation}
{\mathbf{P}_{k|k-1}} =  {\mathbf{F}_k} \cdot  {\mathbf{P}_{k-1|k-1}} \cdot  {\mathbf{F}_k^T} +  {\mathbf{Q}_k}
\end{equation}

\(\mathbf{X}_{k|k-1}\) is the predicted state, \(\mathbf{F}_k\) is the state transition model, and \(\mathbf{P}_{k|k-1}\) is the predicted covariance.

Mathematically, the association can be seen as solving:
\begin{equation}
\arg\min_{\text{associations}} \left( \text{Cost}_{IoU} + \lambda \cdot \text{Cost}_{appearance} \right)
\end{equation}
\(\text{Cost}_{IoU}\) is the cost based on bounding box overlap, \(\text{Cost}_{appearance}\) is the cost based on appearance feature similarity, and \(\lambda\) is a weighting factor.

\subsection{Hungarian Algorithm: Strategy for Tracking}

For targets that are in a "lost" state, we can estimate their potential appearance regions using their past trajectories and additional data such as IMU status. Subsequent frames are then utilized to focus on these predicted regions. We employ the Hungarian algorithm to find the optimal match with the lowest cost. We calculate the distance between the predicted position of each target and the position of the robot detected in the current frame, resulting in a cost matrix. 

If no match is found within a predefined time frame, we consider implementing a binary search. Utilizing the available trajectories and 3D map data, we initiate a binary search from the most recent known position, exploring potential regions until either the target is found or the search boundary is reached.  

\textbf{Our Strategy for Tracking is as follows:} 

\begin{algorithm}
    \DontPrintSemicolon
    \KwData{$D^{YOLOv5}_{current}$, $T^{map}_{tracks}$}
    $C \gets$ ComputeCost($D^{YOLOv5}_{current}$, $T^{map}_{tracks}$)\;
    $Match$, $D_{unmatched}$, $T_{unmatched}$ $\gets$ Hungarian($C$)\;
    \For{each $M$ in $Match$}{
        Update $T^{map}_{tracks}$ with matched $D^{YOLOv5}_{current}$ data\;
    }
    \For{each $unmatched$ in $D_{unmatched}$}{
        Assign ID and tracker for $unmatched$\;
    }
    \For{each $lost$ in $T_{unmatched}$}{
        \eIf{TooLong($lost$)}{
            Remove $lost$ from $T^{map}_{tracks}$\;
        }{
            Update $lost$ status to \textit{MISSING}\;
        }
    }
    \caption{Data Association Strategy}
\end{algorithm}

\subsection{Target Striking}

When targeting, one should seek the optimal solution to the problem, which inherently contains solutions to its sub-problems.

We propose a two-dimensional array structure. The first dimension records the state of the armor when struck, while the second dimension considers factors like the distance from the launcher to the armor plate, the recognized area weight, the projectile launch time, etc., to determine the optimal striking scenario.  

In terms of weighting, it's crucial to assign a higher weight to continuously striking the armor and a secondary weight to the armor plate nearest to the center point. By selecting the armor plate with the highest weight in each step, we ensure a priority in targeting previously hit armor plates and simultaneously eliminate all plates efficiently. 

The coordinate system's center is defined as point (0,0). The provided pseudocode demonstrates our targeting strategy. We employ \textbf{Kalman Filter}  to predict the future location of the armor plate, updating this prediction based on measurements.  

\textbf{\textsc{When we initialize the Kalman Filter, we first define:}} 
\begin{itemize}
    \item  \( \mathbf{X} \): State matrix, describing the current position and velocity of the armor. 
    \begin{align*}
    \mathbf{X} &= 
    \begin{bmatrix}
    0 \\
    0 
    \end{bmatrix}
    \end{align*}
    
    \item  \( \mathbf{F} \): State transition matrix, predicting its position after a duration \( \text{dt} \).
    \begin{align*}
    \mathbf{F} &= 
    \begin{bmatrix}
    1 & \text{dt} \\
    0 & 1 
    \end{bmatrix}
    \end{align*}
    
    \item  \( \mathbf{B} \): Input control matrix, predicting changes to the armor's position due to input conditions.
    \begin{align*}
    \mathbf{B} = 
    \begin{bmatrix}
    \frac{\text{dt}^2}{2} \\
    \text{dt}
    \end{bmatrix}
    \end{align*}

    \item  \( \mathbf{H} \): Measurement mapping matrix, mapping the predicted state space to the actual measurement space.
    \begin{align*}
    \mathbf{H} = 
    \begin{bmatrix}
    1 & 0
    \end{bmatrix}
    \end{align*}

    \item  \( \mathbf{Q} \): Process noise covariance, accounting for ground friction when the chassis is moving, unknown obstacles, and the actual movement of the armor plate might slightly deviate from the prediction. It assists the filter in estimating the deviations.
    \begin{align*}
    \mathbf{Q} = 
    \begin{bmatrix}
    \frac{\text{dt}^4}{4} & \frac{\text{dt}^3}{2} \\
    \frac{\text{dt}^3}{2} & \text{dt}^2 
    \end{bmatrix}
    \cdot \text{std\_acc}^2
    \end{align*}

    \item  \( \mathbf{R} \): Measurement noise covariance, helping us understand the accuracy of our measurements.
    \begin{align*}
    \mathbf{R} = \text{std\_meas}^2
    \end{align*}
    \item  \( \mathbf{P} \): Error covariance matrix, informing us of the estimation accuracy regarding the armor's current position and velocity.
    \begin{align*}
    \mathbf{P} = 
    \begin{bmatrix}
    1 & 0 \\
    0 & 1 
    \end{bmatrix}
    \end{align*}
\end{itemize}
\textbf{\textsc{When Target starts to update, we input these variables into the matrix:}}
\begin{itemize}
    \item The time interval \( \text{dt} \) (corresponding to the computation time interval of YOLOv5).
    \item The system input \( u \) (used to describe explicit motion instructions as a control input).
    \item The standard deviation of acceleration \( \text{std\_acc} \).
    \item The standard deviation of the measurement \( \text{std\_meas} \). 
\end{itemize}


These 9 attributes and parameters collectively enable the Kalman Filter to effectively estimate and track the system's state amidst noisy measurements. In the application of robot armor plate tracking, this is crucial as it allows the system to accurately predict and correct the armor plate's position amidst imperfect measurements and external disturbances. 

\textbf{\textsc{we make them explicit by incorporating the multiplication symbols:}}

\textbf{Predict:}

\begin{equation}
	\mathbf{X} = \mathbf{F} \cdot \mathbf{X} + \mathbf{B} \cdot u 
\end{equation}

\(X\) represents the state of the system. The prediction step initially uses the system's dynamic model to estimate the state of the next time step. This is achieved by multiplying by the transition matrix \(F\). Additionally, if the system is influenced by a known external input \(u\) (np.mean(np.diff($meas[]$)) / $dt$), this influence is also taken into account and is weighted by the matrix \(B\). 

\begin{equation}
	\mathbf{P} = \mathbf{F} \cdot \mathbf{P} \cdot \mathbf{F}^T + \mathbf{Q}
\end{equation}

\( P \) represents the uncertainty or error covariance of the estimated state. This equation describes how to predict the error covariance of the next time step. Among them, \( Q \) is the covariance matrix of the process noise, representing the uncertainty in Kalman. 

\textbf{Update:}

\begin{equation}
	\mathbf{y} = z - \mathbf{H} \cdot \mathbf{X} 
\end{equation}

\( z \) (referred to as $mea[]$) represents the observed values. \( \mathbf{H} \) is the measurement mapping matrix. The residual \( \mathbf{y} \) represents the difference between the actual observed values and the observed values based on predictions.

\begin{equation}
	\mathbf{S} = \mathbf{R} + \mathbf{H} \cdot \mathbf{P} \cdot \mathbf{H}^T 
\end{equation}

it depicts the covariance of measurement prediction error. It consists of two parts: the noise covariance directly from the measurement denoted as \( \mathbf{R} \), and the measurement noise covariance based on the current error covariance estimate \( \mathbf{P} \).

\begin{equation}
	\mathbf{K} = \mathbf{P} \cdot \mathbf{H}^T \cdot \mathbf{S}^{-1} 
\end{equation}

Kalman gain. It determines the level of trust in the new measurement relative to the trust in the current state estimate. A larger value indicates more trust in the new measurement, while a smaller value signifies greater trust in the current estimate. 
\begin{equation}
	\mathbf{X} = \mathbf{X} + \mathbf{K} \cdot \mathbf{y} 
\end{equation}

Update state estimate.\( \mathbf{y} \) is the difference between the actual measurement and the predicted measurement based on the current state estimate.

\begin{equation}
	\mathbf{P} = (\mathbf{I} - \mathbf{K} \cdot \mathbf{H}) \cdot \mathbf{P}
\end{equation}

Update the error covariance. The error covariance describes the uncertainty associated with our state estimate. In this equation, \( \mathbf{I} \) represents the identity matrix. 

\begin{algorithm}
    \DontPrintSemicolon
    \KwResult{$x_{pred}$, $y_{pred}$}
    \SetKwProg{Fn}{Function}{}{end}
    \Fn{\textbf{Kalman\_at\_filter}($x$, $y$)}{
        append $x$ to $meas_x[]$\;
        append $y$ to $meas_y[]$\;
        $dt$ $\gets$ $1 / YOLOv5\_dt + DeepSORT\_dt$\;
        
        $u_x$ $\gets$ np.mean(np.diff($meas_x[]$)) / $dt$\;
        $u_y$ $\gets$ np.mean(np.diff($meas_y[]$)) / $dt$\;

        $std\_acc_x$ $\gets$ np.std(np.diff($meas_x[]$, 2)) / $dt^2$\;
        $std\_acc_y$ $\gets$ np.std(np.diff($meas_y[]$, 2)) / $dt^2$\;

        $std\_meas_x$ $\gets$ np.std($meas_x[]$)\;
        $std\_meas_y$ $\gets$ np.std($meas_y[]$)\;

        $kf^2_x$ $\gets$ Kalman($dt$, $u_x$, $std\_acc_x$, $std\_meas_x$)\;
        $kf^2_y$ $\gets$ Kalman($dt$, $u_y$, $std\_acc_y$, $std\_meas_y$)\;

        \For{each $mx$ in $meas_x[]$}{
            append $kf^2_x$.process($mx$) to $filtered_x$\;
        }
        \For{each $my$ in $meas_y[]$}{
            append $kf^2_y$.process($my$) to $filtered_y$\;
        }

        $x_{pred}$ $\gets$ $filtered_x$[-1][0]\;
        $y_{pred}$ $\gets$ $filtered_y$[-1][0]\;

        \Return $x_{pred}$, $y_{pred}$\;
    }

    \While{True}{
        \For{$i$ in $len(recv)$}{
            $area$ $\gets$ $recv[w] \cdot recv[h]$\;
            $x1$ $\gets$ $recv[x] - middle_x$\;
            $y1$ $\gets$ $middle_y - recv[y]$\;
            $dis\_tmp$ $\gets$ $\sqrt{x1^2 + y1^2}$\;
            \eIf{$area > max\_area$}{
                $max\_area$ $\gets$ $area$\;
            }{}
            \eIf{$dis\_tmp > max\_dis$}{
                $max\_dis$ $\gets$ $dis\_tmp$\;
            }{}
        }

        \For{$i$ in $len(recv)$}{
            $x1$ $\gets$ $recv[x] - middle_x$\;
            $y1$ $\gets$ $middle_y - recv[y]$\;
            $dis\_tmp$ $\gets$ $\sqrt{x1^2 + y1^2}$\;
            $area$ $\gets$ $recv[w] \cdot recv[h]$\;
            $normalized\_dis$ $\gets$ \eIf{$max\_dis == 0$}{inf}{$dis\_tmp / max\_dis$}\;
            $normalized\_area$ $\gets$ $1 - (area / max\_area)$\;
            $weight$ $\gets$ $K\_dis \cdot normalized\_dis + K\_area \cdot normalized\_area$\;
            \If{$weight < min\_weight$}{
                $x\_ret$ $\gets$ $x1 / middle_x$\;
                $y\_ret$ $\gets$ $y1 / middle_y$\;
                $w\_ret$ $\gets$ $recv[w]$\;
                $h\_ret$ $\gets$ $recv[h]$\;
                $min\_weight$ $\gets$ $weight$\;
            }
        }
        $x_{pred}$, $y_{pred}$ $\gets$ Kalman\_at\_filter($x\_ret$, $y\_ret$)\;
    }
    \caption{Target Selection}
\end{algorithm}

\section{Experiment 3: Gimbal Control and Water Pellet Launching Deviation Compensation System}

\subsection{Water pellets fall compensation correction}

The gravitational fall of water pellets is directly proportional to the distance. The effect of the gravitational fall of water pellets on the gimbal movement needs to be considered. 

\definecolor{1_color}{RGB}{128,1,255} 
\definecolor{2_color}{RGB}{255,1,243} 
\definecolor{3_color}{RGB}{4,205,252} 
\definecolor{4_color}{RGB}{255,207,1} 
\definecolor{5_color}{RGB}{37,255,1} 

\begin{table*}
	\small 
	\centering
	\renewcommand{\arraystretch}{1.0}
	\begin{tabular}{|c|c|c|c|c|c|c|c|c|}
		\hline 
		Algorithms & MSE & RMSE & $R^2$ \\
		\hline
		\textbf{\textcolor{1_color}{Support Vector Regression}} & 0.0230712 & 0.1518921 & 0.9828452 \\
		\textbf{\textcolor{5_color}{K-Nearest Neighbors Algorithm}}* & \textbf{0.0164379} & \textbf{0.1282102} & \textbf{0.9869846} \\
		\textbf{\textcolor{3_color}{4th-Degree Polynomial Regression}} & 0.0234895 & 0.1532628 & 0.9825341 \\
		\textbf{\textcolor{4_color}{5th-Degree Polynomial Regression}} & 0.0231119 & 0.1520259 & 0.9828149 \\
		\hline
	\end{tabular}
	\caption{\centering Presents the evaluation metrics (MSE, RMSE, $R^2$) for different algorithms, highlighting their effectiveness in data fitting.}
	\label{tab:fall-algorithm}
\end{table*}

The YOLOv5 algorithm has detected the pixel value \( h_{\text{CNN}} \) corresponding to the armor in the current image. Based on this pixel value, combined with camera parameters and IMU data, we can derive formulas for object detection and distance estimation. 

Given the pixel height of the object as \( h_{\text{CNN}} \), the real height of the object as \( H_{\text{r}} \), and the focal length of the camera as \( f \), the estimated distance \( D \) between the object and the camera can be derived as: 
\begin{equation}
D = \frac{f \times H_{\text{r}}}{h_{\text{CNN}}}
\end{equation}

Assuming the IMU provides the rotational angles of the gimbal as \( \text{roll} \), \( \text{pitch} \), and \( \text{yaw} \), we can determine the rotation matrix \( R \) of the gimbal. For a point \( P_{\text{camera}} = (x, y, z) \) in the camera's coordinate system, its position \( P_{\text{r}} \) in the real coordinate system can be computed as: 
\begin{equation}
P_{\text{r}} = R \times P_{\text{camera}}
\end{equation}

Taking into account the data from both the IMU and the camera, the three-dimensional coordinates of the object \( P_{\text{r}} \) can be obtained through: 
\begin{equation}
P_{\text{r}} = R \times \left( \frac{x_{CNN} \times D}{f}, \frac{y_{CNN}\times D}{f}, D \right)
\end{equation}

While we adapted the actual drop data of the water pellet to the degree of closeness calculation, it's essential to note that the camera and the water pellet's launch position do not reside on the same plane or point. Instead, the camera is positioned 10cm forward and 6cm upward from the water pellet launch location. Consequently, traditional physical models are not applicable to this design. Thus, we fit the actual falling data of the water pellet with the closeness calculation.

\textbf{\textsc{The methodology for fitting employed in this research encompasses the following steps:}} 1) An infrared distance sensor is placed at a fixed position on the turret. 2) The armor plate hit status is checked. If it is\_hit, the parameter y from the camera, the value from the infrared distance sensor are recorded. 3) The above actions are repeated from the farthest detection distance to the turret lowest mechanical limit. 

For our experiments, we utilized a camera with specifications: 1/2.7inch OV2710, resolution of 640x480, frame rate of 120fps, focal length of F.3.6mm, FOV of 90°. The central axis of the camera aligns with the centerline of the water pellet launch. \textit{Dropping Distance} represents the deviation value from the center point \textit{(0,0) center position} of the armor plate recognized by YOLOv5, with the unit being pixels (px). $D(\mathbf{Camera}, \mathbf{Armor})$ indicates the distance from the camera's CMOS to the armor plate, measured in centimeters (cm). 

\begin{figure*}[t]
	\centering
	\captionsetup{justification=centering} 
	\includegraphics[width=0.8\textwidth]{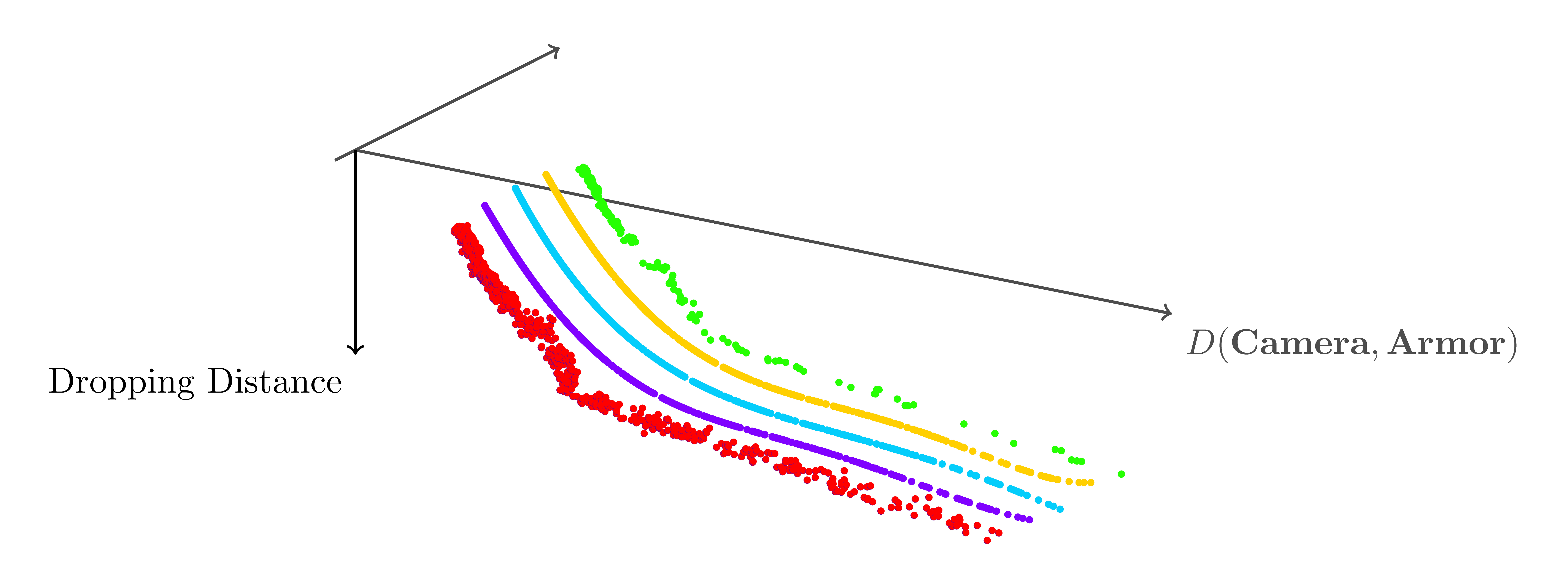}
	\caption{Regression parameters for each algorithm. \textcolor{red}{Red} represents the actual parameters of the water pellet falling.}
	\label{fig:task-sequence}
\end{figure*}

We introduced a coordinate system wherein the x-coordinate represents $D(\mathbf{Camera}, \mathbf{Armor})$ and the y-coordinate corresponds to the \textit{Dropping Distance}. We used image fitting to maximize the fit for the scattered data points. Several regression algorithms were used for the fitting calculation:

\begin{enumerate}
	\item \textbf{Polynomial Regression, PR*}
	
	Polynomial Regression tries to fit the data to a polynomial. For instance, the formula for a quadratic polynomial regression is:
	\begin{equation}
		y = \beta_0 + \beta_1 x + \beta_2 x^2 + … + \beta_{n-1} x^{n-1} + \beta_n x^n +\epsilon
	\end{equation}
	\(y\) is the target variable, \(x\) is the feature, \(\beta_0, \beta_1,\) and \(\beta_2\) are the model parameters, and \(\epsilon\) is the error term.
	
	\item \textbf{Support Vector Regression, SVR}
	
	Support Vector Regression tries to find a function that keeps the error of most data points within a predetermined range. The formula is more complex, but in its basic form:
	\begin{equation}
		f(x) = \sum_{i=1}^{N} (\alpha_i - \alpha_i^*) K(x, x_i) + b
	\end{equation}
	\(\alpha_i\) and \(\alpha_i^*\) are Lagrange multipliers, \(K\) is a kernel function (like linear, radial basis, etc.), and \(x_i\) are the support vectors.
	
	\item \textbf{K-Nearest Neighbors Algorithm, KNN}
	
	KNN based on the distances in its training data. For a new input sample, KNN will find the \(K\) nearest points in the training set and make predictions based on the average or majority voting principle of these points. The algorithm can be represented as:
	\begin{equation}
		y = \frac{1}{K} \sum_{i=1}^{K} y_{(i)}
	\end{equation}
	\(y_{(i)}\) is the target value of the \(i\)-th nearest neighbor.
\end{enumerate}

We use Mean Squared Error (MSE) and Root Mean Squared Error (RMSE) to evaluate the performance of the regression curve:  

\begin{equation}
\text{MSE} = \frac{1}{n} \sum_{i=1}^{n} (y_i - \hat{y}_i)^2
\end{equation}

\begin{equation}
\text{RMSE} = \sqrt{\frac{1}{n} \sum_{i=1}^{n} (y_i - \hat{y}_i)^2}
\end{equation}

$y_i$ is the actual observation, $\hat{y}_i$ is the predicted value, and $n$ is the number of observations. \newline

Mean Absolute Error (MAE): This is the average of the differences between actual and predicted values.
\begin{equation}
\text{MAE} = \frac{1}{n} \sum_{i=1}^{n} |y_i - \hat{y}_i|
\end{equation}

The coefficient of determination, often denoted as \( R^2 \), is used to quantify the proportion of the variance in the dependent variable that is predictable from the independent variable(s). It is given by: 

\begin{equation}
R^2 = 1 - \frac{\text{SSE}}{\text{SST}}
\end{equation}

\begin{itemize}
    \item \text{SSE} is the sum of squares of the residuals, given by 
    \begin{equation}
        \sum_{i=1}^{n} (y_i - \hat{y}_i)^2
    \end{equation}
    \item \text{SST} is the total sum of squares, given by 
    \begin{equation}
        \sum_{i=1}^{n} (y_i - \bar{y})^2
    \end{equation}
    with \( \bar{y} \) being the mean of the observed data.
\end{itemize}

In (Table \ref{tab:fall-algorithm}), We present the performance data of MSE, RMSE, and $R^2$ within different algorithms.

\subsection{Removing gimbal response noise}
During the actual target tracking process, there may be delays and coordination issues between the gimbal and the vision system. Therefore, we employed a \textit{FIR (Finite Impulse Response)} to control the movement of the gimbal. FIR filters can be designed directly for given frequency characteristics and implemented in a non-recursive manner. \textbf{\textsc{FIR systems offer several distinct advantages:}} 1) The system is always stable, 2) it is easy to achieve precise linear phase characteristics, 3) it allows the design of multi-band (or multi-stopband) filters. 

\begin{figure*}[t]
	\centering
	\captionsetup{justification=centering} 
	\includegraphics[width=0.8\textwidth]{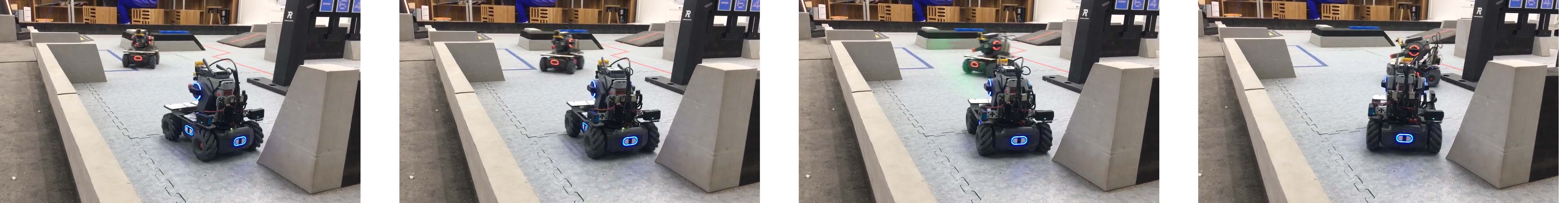}
	\caption{Reflects the aiming status of the Automatic Assisted Aiming System, A sequence of four images representing the progression of states over time, from left (earliest) to right (latest).}
	\label{fig:RM_move}
\end{figure*}

FIR filter is defined as follows: 

Assume an FIR filter has an order of \( N \), and its unit impulse response is \( h[n] \). Given an input signal \( x[n] \), the output signal \( y[n] \) can be expressed as: 

\begin{equation}
y[n] = \sum_{k=0}^{N} h[k] \cdot x[n-k]
\end{equation}

\begin{itemize}
    \item \( y[n] \) is the value of the output signal at time \( n \). 
    \item \( h[k] \) represents the coefficients of the FIR filter, also known as the value of the unit impulse response at time \( k \). 
    \item \( x[n-k] \) is the value of the input signal at time \( n-k \).
    \item \( N \) represents the order of the filter, which is equivalent to the number of filter coefficients minus one. 
\end{itemize}

It illustrates the working principle of an FIR filter. Each output value is a weighted sum of the current and preceding input values, multiplied by the filter coefficients. 

\subsection{Incremental PID Controller with Feedforward Enhancement}

After achieving stable control of the gimbal and determining the actual impact point, we employ a PID controller with feedforward enhancement for precise control. The controller's primary aim is to compute the output based on the given setpoint and the actual process value. We have introduced a rate limit and an enhancement threshold for improved performance. 

Unlike traditional PID controllers that determine their output based on the present error value, Incremental PID controllers emphasize the error's change from one time step to the next. This focus offers an advantage, especially in systems with parameters that drift over time, ensuring effectiveness as the system evolves. Incremental PIDs are also less susceptible to sudden error spikes, enhancing stability in specific scenarios.


Incremental PID controllers operate based on error changes rather than the error itself, leading to more responsive corrections as they adjust to new disturbances. Coupled with the feedforward enhancement, which anticipates setpoint changes, the control action becomes smoother. Together, they allow for faster responses without amplifying noise and ensure stability in dynamic situations. 

The rate limit ensures that the control output doesn't change too rapidly, which can prevent oscillations or instability in the controlled system. The enhancement threshold can be set to determine when the feedforward enhancement should be applied, ensuring that it's used only when beneficial for the control action. 

However, caution is necessary when implementing this controller. The derivative term in PID controllers can amplify high-frequency noise, making a robust noise filtering mechanism essential. Moreover, determining the correct parameters \( K_p, K_i, K_d, \) and \( K_f \) can be challenging, particularly in systems with intricate dynamics. 

This controller is particularly useful for systems that experience frequent setpoint changes or demand rapid disturbance responses. Its main advantage lies in its swift adaptation to system dynamics, reducing overshoots and enhancing settling times. 

\begin{itemize}
    \item The error \( e(t) \) is defined as the discrepancy between the current setpoint and the actual reading.

    \item The incremental output, which embeds both the PID control and Feedforward Enhancement, is formulated as:

    \begin{equation}
    \Delta u(t) = K_p \Delta e(t) + K_i \int e(t) dt + K_d \frac{d e(t)}{dt} + K_f \frac{\Delta F(t)}{\Delta t}
    \end{equation}

    \begin{itemize}
        \item \( \Delta e(t) \) represents the difference between the current and previous error.
        \item \( \Delta F(t) \) signifies the second difference of the feedforward term.
    \end{itemize}

    \item To ensure stability, the output is circumscribed by a rate limit, maintaining it within a predefined range.
\end{itemize}

\section*{Conclusion and Discussion}
In crucial moments of the Robomaster Youth Championship, the Robomaster EP robot utilizes autonomous algorithms to maintain a competitive edge. This system is expected to provide a solid technical foundation for the Robomaster EP robot in future tournaments, with its most notable feature being its enhanced accuracy in target engagement. 

Before 2021, there was a noticeable gap in research related to auto-aiming system design for Robomaster EP based on unofficial modules. This study aims to fill that void. 

We have successfully implemented tracking, identification, prediction, and target engagement for the Robomaster EP robot. However, during the competition, we did not conduct precise calculations concerning data latency errors. Moving forward, we will focus on improving the latency compensation system and are enthusiastic about incorporating human posture detection to refine the robot's specific actions further.

\section*{Acknowledgments}

\quad The initial findings of this paper emerged in November 2021. After numerous technical iterations, we finalized and published the paper in October 2023. 

We would like to extend our gratitude to The High School Affiliated to Minzu University of China for providing us with the environment and platform to complete our system, allowing us to showcase our talents here. 

We extend our heartfelt gratitude to \textbf{Professor Jianli Yang, Professor Shibin Zhang} for their invaluable guidance throughout the development of this paper. Their insights into the paper writing process and methodologies greatly enriched our understanding, paving the way for the successful completion of this work. While a single paper cannot encapsulate our entire journey in technological development, we are inspired by both professors' unwavering commitment to continuous learning and a diligent, truth-seeking research approach.  

Our heartfelt appreciation goes to \textbf{All Members of The HAOYE Team, The High School Affiliated to Minzu University of China}. Their rigorous training and hands-on exercises helped us pinpoint various system challenges. Their recurrent feedback for system enhancements, coupled with the provision of relevant materials and paper-writing techniques, have been of immense help. We are profoundly touched by their spirit of camaraderie and collaboration.  

We are grateful for the generous support provided by Beijing HISINGY. 

Their contributions have been invaluable to our paper. We are deeply moved by their unwavering support, and the lessons learned will stay with us throughout our journey. With the knowledge acquired, we are poised to make impactful contributions in our field of expertise.

\bibliographystyle{plain}
\nocite{*}
\bibliography{references}

\clearpage
\appendix

\end{document}